\documentclass[a4paper,11pt]{article}

\usepackage{fourier}
\usepackage{color}
 \usepackage{graphicx}
\usepackage{url}
\usepackage[affil-it]{authblk}
\usepackage{amsmath}
\usepackage{wrapfig}
\usepackage{booktabs}
\usepackage[T1]{fontenc}
\usepackage{times}

\usepackage{subfigure}
\usepackage{subcaption}
\usepackage{graphicx}

\begin{document}

\title{Subgraph Clustering and Atom Learning\\for Improved Image Classification}

\author{Aryan Singh}
\author{Pepijn Van de Ven}
\author{Ciarán Eising}
\author{Patrick Denny}
\affil{University of Limerick}
\date{}
\maketitle
\thispagestyle{empty}

\noindent\textbf{Preprint Notice:}

\noindent This paper is a preprint of a paper submitted to the 26th Irish Machine Vision and Image Processing Conference (IMVIP 2024). If accepted, the copy of record will be available at IET Digital Library. 

\begin{abstract}

In this study, we present the Graph Sub-Graph Network (GSN), a novel hybrid image classification model merging the strengths of Convolutional Neural Networks (CNNs) for feature extraction and Graph Neural Networks (GNNs) for structural modeling. GSN employs k-means clustering to group graph nodes into clusters, facilitating the creation of subgraphs. These subgraphs are then utilized to learn representative `atoms` for dictionary learning, enabling the identification of sparse, class-distinguishable features. This integrated approach is particularly relevant in domains like medical imaging, where discerning subtle feature differences is crucial for accurate classification.

To evaluate the performance of our proposed GSN, we conducted experiments on benchmark datasets, including PascalVOC and HAM10000. Our results demonstrate the efficacy of our model in optimizing dictionary configurations across varied classes, which contributes to its effectiveness in medical classification tasks. This performance enhancement is primarily attributed to the integration of CNNs, GNNs, and graph learning techniques, which collectively improve the handling of datasets with limited labeled examples. Specifically, our experiments show that the model achieves a higher accuracy on benchmark datasets such as Pascal VOC and HAM10000 compared to conventional CNN approaches.
\end{abstract}

\section{Introduction}
CNNs have transformed image classification, showcasing impressive performance by learning hierarchical feature representations through layers of convolutions and pooling. Nevertheless, traditional CNNs may face challenges in specific contexts, particularly when dealing with datasets characterized by significant class overlap or sparsely labeled data, as often encountered in medical imaging \cite{DINSDALE20223866}.

A significant issue arises from the utilization of pre-trained CNNs, which are typically trained on extensive datasets like ImageNet, having diverse classes \cite{imagenet}. While these models excel in generalization, they may encounter difficulties in scenarios requiring finer distinctions or handling multi-label data. In such cases, the learned feature maps might exhibit overlap or intertwining between representations of distinct objects, leading to diminished classification accuracy, a concern heightened in medical imaging due to the critical diagnostic implications of subtle feature variations \cite{cnn_med_lim}.

To address these limitations, there has been a burgeoning interest in amalgamating dictionary learning techniques with the feature extraction capabilities of Deep Neural Networks (DNNs), particularly CNNs. Dictionary learning aims to identify a set of representative elements, or `atoms`, which, when combined, can accurately reconstruct the original data. This set is known as the 'dictionary'\cite{deep_dict}. By learning a dictionary tailored to the specific dataset, the model can more effectively isolate and represent the diverse features present in images. This enhanced representation allows the model to capture subtle nuances and variations that might be missed by other feature extraction methods, leading to improved classification performance \cite{dict_graph,dict1}. 

Our research extends upon these foundational concepts, advancing the integration of CNNs, dictionary learning, and GNNs within the innovative framework of Graph Sub-Graph Network (GSN). We propose a novel image classification model tailored for scenarios characterized by multi-label classes and sparsely labeled data. Our approach harnesses pre-trained CNN features and segment nodes (image regions) based on similarity metrics and learns `atoms` from these clusters. These atoms collectively form our dictionary, resulting in superior classification performance compared to traditional DNNs, particularly in the way the model's performance decline is less pronounced as training data is reduced. 

\begin{figure}[h]
\centering
\subfigure[HAM10000 dataset (7 classes)]{
  \includegraphics[width=0.45\textwidth,height=6cm]{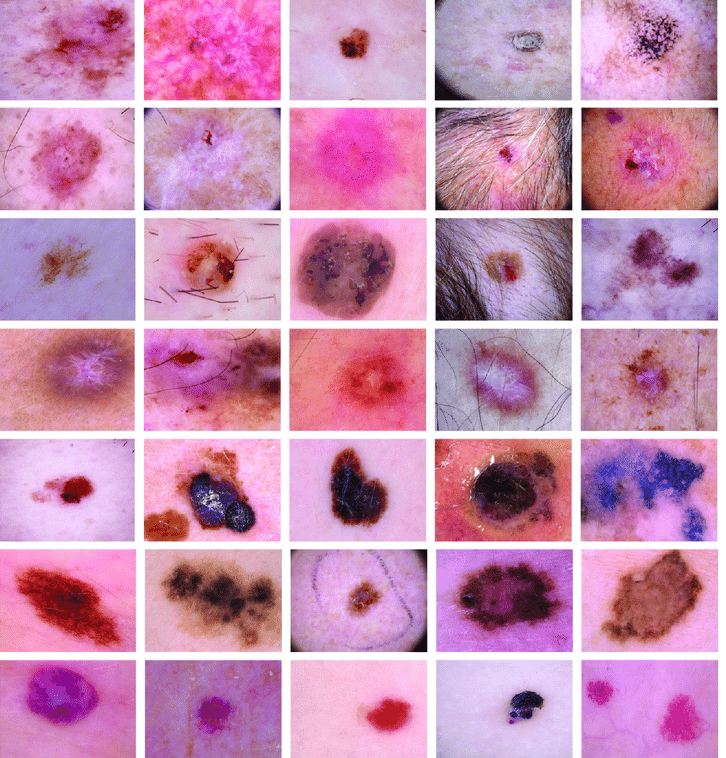}
  \label{fig:ham10000}
}
\hspace{3mm}
\subfigure[Pascal VOC dataset (20 classes)]{
  \includegraphics[width=0.45\textwidth,height=6cm]{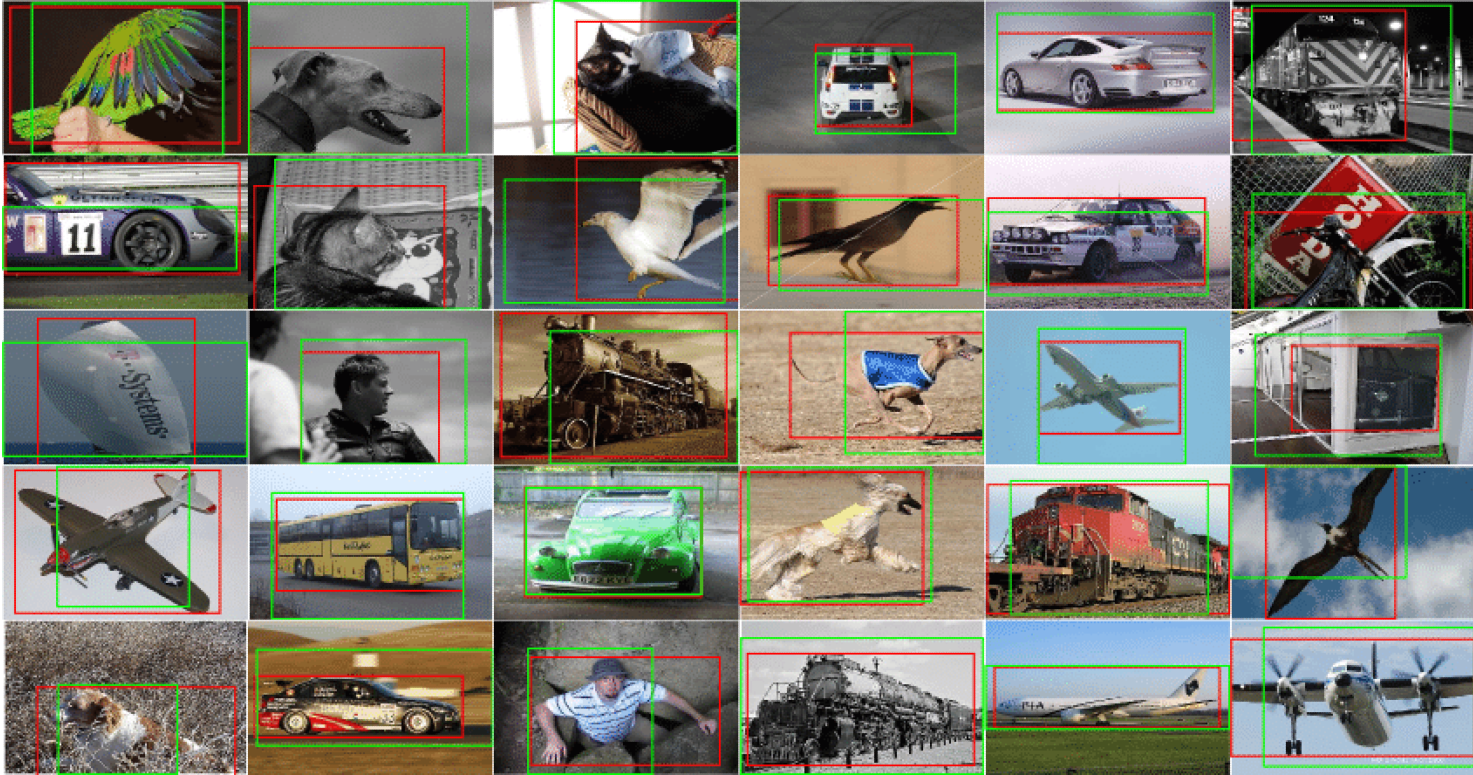}
  \label{fig:pascalvoc}
}
\caption{Datasets used in this study. (a) shows the HAM10000 dataset with 7 classes, while (b) presents the Pascal VOC dataset with 20 classes.}
\label{fig:datasets}
\end{figure}


\section{Prior Art}
The field of computer vision has experienced a profound transformation driven by the remarkable advancements in CNNs and DNNs. Models such as ResNet18 have achieved groundbreaking success in image classification, demonstrating the immense potential of deep learning for visual understanding. Deep Neural Networks (using convolutions or transformers) have become the dominant paradigm in computer vision, enabling solutions to a wide array of challenging problems. Their impact extends far beyond generic computer vision tasks, revolutionizing the field of medical imaging. CNNs and DNNs are now instrumental in tasks like medical image analysis, disease detection, and treatment planning. Their ability to learn complex patterns from medical scans has led to significant improvements in accuracy, efficiency, and early diagnosis in various medical domains. This section explores work done in the field of computer vision using DNNs and Dictionary learning. 

\subsection{CNN}
CNNs\cite{cnn} get their name from the `convolution` operation, a core building block of these networks. Imagine a small window (called a kernel) sliding over an image, focusing on a small section at a time. At each position, the kernel multiplies its own values with the corresponding image pixels and sums up the results. This creates a single number for that specific location, which gets stored in a new output map. The key goal of convolution is to take image data and gradually distill it down, helping the network understand larger-scale features and patterns within the image. Convolution works like a series of intelligent filters that analyze an image at multiple levels of detail. Early on, it identifies simple patterns like edges and lines based on how neighboring pixels relate. As the process goes deeper into the network, the filters build upon these basic patterns, detecting more complex shapes and objects. The result is a layered understanding of what is in the image, from small details to overall concepts. Finally, all this rich information is condensed into a streamlined form. This compressed version can then be used for all sorts of tasks, like deciding what the image shows, pinpointing objects, or highlighting distinct areas.


\textbf{ResNet}  \cite{resnet} is a breakthrough deep neural network architecture designed to overcome the challenges of training extremely deep networks\cite{vanexp}. Unlike traditional networks, ResNets introduce `residual blocks`. These blocks use skip connections \cite{skipnet} to add the original input directly to the output of a group of convolutional layers. This helps the network focus on learning the changes between the input and desired output, rather than trying to re-learn features from scratch.

ResNet achieved groundbreaking success on the ImageNet dataset \cite{imagenet}, significantly advancing the state-of-the-art in image classification at the time.  Its innovative architecture has become the building block of numerous computer vision tasks, demonstrating its effectiveness as a powerful and adaptable tool.

ResNet's abilities have proven invaluable in medical image analysis. For example, it has been used to successfully detect skin cancer from images. Additionally, ResNet architectures have aided in the diagnosis of brain diseases using MRI scans  \cite{Lu2020}, and in the segmentation of lung tumors in CT images \cite{lungtumor}. These examples highlight ResNet's versatility in extracting complex patterns from medical images, contributing to improved diagnostic tools.

However, there are limitations in transferring knowledge directly from a large dataset like ImageNet to medical image analysis tasks. While pre-trained ResNets offer a strong foundation, medical images often contain unique features and subtle class distinctions compared to natural images. This can lead to the model overfitting to the general features learned on ImageNet, hindering its ability to differentiate between fine-grained medical classes. 

These limitations have led to the exploration of advanced techniques to further enhance the performance of deep learning models in medical image analysis \cite{cnn_med}. Attention models have proven particularly effective, allowing models to focus on the most relevant regions of an image. Additionally, autoencoders can be used to learn compact, informative representations tailored to specific medical imaging tasks.  Another promising approach to address these challenges is that of dictionary learning in computer vision, which offers new solutions for extracting and representing the subtle, distinguishing features crucial in medical image analysis.

\subsection{Dictionary Learning}
Dictionary learning is a powerful technique in representational learning aimed at deriving a set of basis elements, known as `atoms` which can reconstruct signals or images via sparse combinations \cite{dict_start1}. This approach provides a more flexible and richer representation than traditional bases like Fourier or wavelet transforms. Its applications include the reduction of noise in images, the filling in of missing image parts, and the enhancement of image resolution \cite{dict_super,dict_dnos}.

Given a set of training samples \( Y = [\mathbf{y}_1, \mathbf{y}_2, \dots, \mathbf{y}_N] \in \mathbb{R}^{n \times N} \), where each \(\mathbf{y}_i\) is an \(n\)-dimensional column vector representing a sample, the goal of dictionary learning is to learn a dictionary matrix \( D = [\mathbf{d}_1, \dots, \mathbf{d}_k] \in \mathbb{R}^{n \times k} \) and a sparse coefficient matrix \( X = [\boldsymbol{\alpha}_1, \dots, \boldsymbol{\alpha}_N] \in \mathbb{R}^{k \times N} \) such that the training samples \( Y \) can be approximated as a linear combination of the atoms in \( D \) with coefficients in \( X \):

\begin{equation}
\label{eq:dict}
Y \approx D X^T
\end{equation}
The dictionary learning process involves optimizing the following objective function, subject to the specified constraints:

\begin{equation}
\min_{D, X} \left\| Y - DX^T \right\|_F^2 + \lambda \left\| X \right\|_1 \quad \text{subject to} \quad \left\| d_k \right\|_2 = 1, \text{ for } k=1,\ldots,K
\end{equation}

where:
\(\left\| Y - DX^T \right\|_F^2\) is the Frobenius norm (squared) of the reconstruction error, measuring how well the dictionary and sparse coefficients approximate the original data, \(\left\| X \right\|_1\) is the \(\ell_1\) norm of the sparse coefficient matrix, promoting sparsity in the representation, \(\lambda\) is a regularization parameter that controls the trade-off between reconstruction accuracy and sparsity, \(\left\| d_k \right\|_2 = 1\) is the constraint that enforces unit norm on each atom in the dictionary.



Dictionary learning's capability to capture essential image patterns and textures underpins its success in computer vision, making it a critical tool for extracting and representing subtle, distinguishing features in complex datasets.

\subsubsection{Dictionary Learning for Non-Euclidean Data}

Traditional signal processing and dictionary learning techniques have primarily been applied to data situated within Euclidean spaces. However, the growing ubiquity of data embedded in complex structures like graphs—including those representing social networks, sensor networks, and biological interaction networks demands a paradigm shift. Unlike signals in Euclidean spaces, the structure and topology of graphs deeply influence the support of graph-based signals, revealing crucial information that extends beyond mere signal values.

Dictionary learning has thus become a pivotal technique for achieving sparse representations of signals in non-Euclidean spaces. This approach extensively utilizes the graph Laplacian, a fundamental operator in spectral graph theory, to tailor dictionaries to the unique spectral characteristics of graph signals. The graph Laplacian \(\mathbf{L}\) is defined as \(\mathbf{L} = \mathbf{D} - \mathbf{A}\), where \(\mathbf{A}\) is the adjacency matrix of the graph and \(\mathbf{D}\) is the degree matrix, a diagonal matrix where each element \(D_{ii}\) equals the sum of the edges connected to vertex \(i\), facilitating an analysis that incorporates the graph's inherent structure \cite{dict_graph}.

These learned dictionaries enable the efficient approximation of graph signals using only a few significant components. This sparsity is vital for effective analysis, compression, and understanding of data in these complex domains.

The adaptation of dictionary learning for graph-based signals underscores a significant evolution in signal processing. It demonstrates the importance of modifying classical methods to handle the diverse, non-Euclidean structures found in modern data \cite{gra_dict_lr}. These advancements pave the way for analyzing and extracting meaningful information from the increasingly intricate datasets encountered across a wide range of fields.


\subsection{GNN}
GNNs \cite{gnn} are specialized neural networks designed for processing data that can be represented as graphs. These networks handle entities (nodes) and their connections (edges), capturing significant structural and relational information. GNNs operate based on message passing, aggregation, feature updates, and forward propagation.

\begin{itemize}
    \item \textbf{Message Passing:} Each node sends and receives messages based on its own features and those of its adjacent nodes \cite{mpass}. This process allows information to flow through the graph, enabling nodes to gather insights from their local neighborhoods.
    \item \textbf{Aggregation:} Nodes aggregate the messages received from their neighbors using functions such as sum, mean, or max, forming a single representative vector. This synthesis is crucial for integrating information from multiple sources within the graph.
    \item \textbf{Feature Update:} Nodes update their own state by transforming the aggregated vector through a neural network function, integrating the node’s existing state with the newly aggregated information.
    \item \textbf{Forward Propagation:} The updated node features are then processed through additional network layers, transforming the features into higher-level representations. This step mirrors operations in conventional neural networks and is essential for feature learning.
\end{itemize}

The general update rule for a node \( v \) in a GNN is given by the equation:
\begin{equation}
h_v^{(l+1)} = f\left(h_v^{(l)}, \bigoplus_{u \in \mathcal{N}(v)} g(h_u^{(l)}, h_v^{(l)}, e_{uv})\right),
\end{equation}
where \( h_v^{(l)} \) is the node feature vector at layer \( l \), \( \mathcal{N}(v) \) denotes the set of neighboring nodes, \( e_{uv} \) are the edge features, \( f \) and \( g \) represent learnable neural network functions, and \( \bigoplus \) is the aggregation function.

Within this framework, Graph Convolutional Networks (GCNs) extend the concept of convolution from regular grid data, such as images, to graph-structured data \cite{gcnn}. They leverage the spectral properties of graphs, utilizing the eigendecomposition of the graph Laplacian $\mathbf{L}$. The convolution operation in a GCN is defined as:
\begin{equation}
H^{(l+1)} = \sigma\left(\mathbf{U} \mathbf{\Lambda} \mathbf{U}^T H^{(l)} W^{(l)}\right),
\end{equation}
where \( H^{(l)} \) are the node features at layer \( l \), \( \mathbf{U} \) contains the eigenvectors of \( \mathbf{L} \), \( \mathbf{\Lambda} \) is the diagonal matrix of eigenvalues, \( W^{(l)} \) is the weight matrix for that layer, and \( \sigma \) is a nonlinear activation function. Through these mechanisms, GCNs can effectively capture and analyze the intricate relationships and structural nuances present in graph data.

\section{Methodology}
At the core of our approach lies the integration of dictionary learning, GNNs, and concepts from computer vision to achieve optimal class representation. Our method integrates a series of components designed to optimally exploit the structural nuances of graph data, including node selection, subgraph creation, embedding generation, and classification. In our study, we have utilized features extracted from a DNN and applied dictionary learning techniques to learn sparse representations using GNN for optimal class representation. 

This section elaborates on the methodology employed to represent and analyze images through advanced graph-based techniques.

\subsection{Graph Construction}
\label{subsec:gconst}
The initial step in our methodology involves constructing a graph representation of the image, which begins by segmenting the image into superpixels. These superpixels serve as nodes within the graph. We extract features from these nodes using a pre-trained DNN, specifically frozen ResNet18 in our experiments. The extracted features are high-level descriptors of the visual content within each superpixel, capturing detailed texture, color, and structural information.

Once the nodes are defined with their corresponding features, we proceed to establish edges between them. This is achieved through a K-nearest neighbors (KNN) approach, where cosine similarity serves as the metric to determine the connections between nodes. To leverage the grid structure of the image we have also assign 4 adjacent nodes as neighbours. This graph structure highlights the intricate interactions across the image, facilitating a deeper understanding of the spatial and feature-based relationships within the image. 


\subsection{Graph Component Clustering}
Following the graph construction, K-means clustering is employed to partition the graph into several clusters. Each cluster groups nodes with similar features, likely representing distinct patterns or regions within the image. This step is crucial for identifying key areas of interest that are significant to the subsequent classification task.

\subsection{Subgraph Generation and Atom Learning}
Each cluster identified from the previous step is then used to generate a subgraph. These subgraphs are processed using a GCN. The GCN learns an embedding, termed as an `atom` which encapsulates the spectral properties and relational dynamics specific to the subgraph. These atoms provide a compact yet comprehensive representation of distinct regions within the image.


\subsection{Dictionary Construction}
A dictionary of visual patterns is constructed by aggregating the atoms learned from each significant graph cluster. These atoms serve as fundamental building blocks for representing and understanding the image content. Each atom embodies a compact yet informative embedding derived from a specific subgraph, encapsulating the unique spectral and relational characteristics of the corresponding image region.

The dictionary serves as a repository of diverse visual primitives, akin to a visual vocabulary. By assembling atoms from various regions of the image, the dictionary captures a wide range of local patterns and structures. This rich collection of atoms enables the subsequent classification process to accurately and efficiently characterize new images by identifying and quantifying the presence of these learned patterns.

\subsection{Classification by Feature Concatenation}
Once the dictionary is constructed, the final classification is performed by concatenating the features corresponding to different atoms. This concatenated features is then fed into a classifier to assign the image to one of the predefined categories.

\begin{figure}
 \centering
 \includegraphics[width=1\textwidth,height = 6cm]{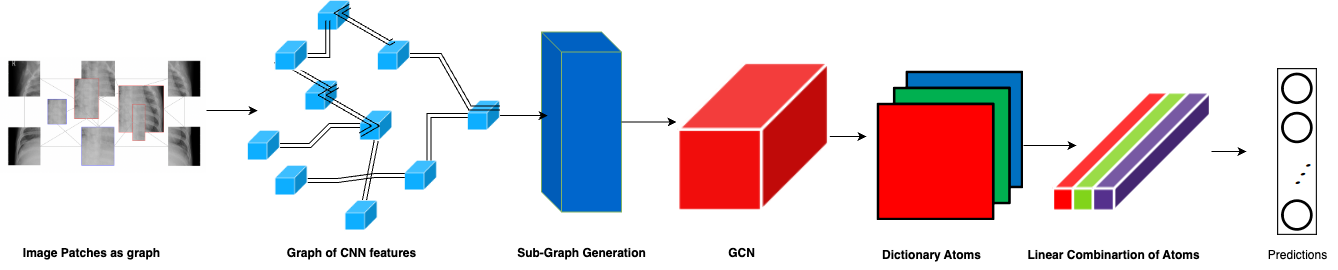}
 \vspace{-4mm}
 \caption{
  Graph Sub-Graph Network model architecture.
 }
 \label{fig:proposed}
 \vspace{-0.5cm}
\end{figure}

\section{Results}
We conducted an evaluation of our proposed model on two benchmark datasets that are highly regarded in the image classification domain. Pascal VOC, comprising 20 diverse classes \cite{pascal}, and HAM10000, which includes 7 skin cancer classes \cite{ham}. Our method is implemented using PyTorch and PyTorch Geometric for graph convolution operations.  Model training is conducted with an initial learning rate of 0.001, employing ReduceLROnPlateau to dynamically adjust the learning rate during training for a maximum of 100 epochs.

On the Pascal VOC dataset, our method achieves a peak accuracy of 88.63\%. This is a substantial improvement over the baseline ResNet-18 model, which achieves an accuracy of 63\%. Similarly, on the HAM10000 dataset, our model exceeds the performance of popular ResNet models. Table \ref{tab:tab1}, shows our GSN model's accuracy of 89.63\% is slightly higher than the closest ResNet variant. By utilizing a frozen ResNet18 model's (which does not require training) features as explained in Section \ref{subsec:gconst}, our approach excels across two diverse benchmark datasets.

\begin{wraptable}{l}{0.5\textwidth} 
\caption{Model Accuracy}
\label{tab:tab1}
\begin{tabular}{@{}lcc@{}}
\toprule
Model (Parameters)  & Pascal VOC & HAM10000 \\ \midrule
ResNet-18 (11.19M)  & 63.43\%    & 88.85\%  \\
ResNet-34 (21.29M) & 66.03\%     & 84.54\% \\
ResNet-50 (23.55M)  & 84.01\%    & 88.76\%  \\
\bf{GSN} (50.57K)        & \textbf{88.63}\%    & \textbf{89.63}\%  \\ \bottomrule
\end{tabular}
\vspace{0pt}
\end{wraptable}

\section{Conclusion}
In this paper, we introduce the GSN, a novel image classification framework that reimagines the integration of dictionary learning and GNNs. GSN distinguishes itself by leveraging graph representations to capture the intricate relationships between image features, combined with a sparse coding approach for optimal class representation. Our experiments demonstrate that GSN, leveraging features extracted from ResNet18, not only achieves competitive performance on standard benchmarks like Pascal VOC but also surpasses the performance of ResNet18, ResNet34, and ResNet50 on the challenging HAM10000 medical dataset, showcasing the effectiveness of our graph-based approach in utilizing these features. This success is achieved with a significantly reduced parameter count compared to traditional DNNs, highlights the potential of GSN to deliver both accuracy and efficiency in image classification tasks. The unique fusion of graph-based learning and sparse coding in GSN opens up new avenues for exploring the power of structural representations in image understanding. Future research will focus on extending GSN to a broader range of applications, refining graph construction techniques, and delving deeper into the interpretability of the learned graph representations.

\appendix

\bibliographystyle{apalike}

\bibliography{imvip}

\end{document}